\documentclass[conference,10pt,letter]{IEEEtran}
\usepackage[latin1]{inputenc}
\usepackage{graphicx}
\usepackage{stmaryrd}

 \newcommand{\eR}{\mbox{I\hspace{-.15em}R}}

\begin{document}

\title{Inclusion within Continuous Belief Functions}

\author{\authorblockN{Dorra Attiaoui}
\authorblockA{University of Tunis\\
       LARODEC, ISG Tunis,\\
       Tunisia\\
       Email: attiaoui.dorra@gmail.com}
\and
\authorblockN{Pierre-Emmanuel Dor\'e}
\authorblockA{University of Rennes 1\\
Rue \'Edouard Branly BP 30219\\
22302 Lannion, France\\
Email: pierre-emmanuel.dore@univ-rennes1.fr}
\and
\authorblockN{Arnaud Martin}
\authorblockA{IRISA\\
University of Rennes 1\\
Rue \'Edouard Branly BP 30219\\
22302 Lannion, France\\
Email: Arnaud.Martin@univ-rennes1.fr}
\and
\authorblockN{Boutheina Ben Yaghlane\\
\authorblockA{University of Carthage,\\
       LARODEC, IHEC Carthage,\\
       Tunisia}\\
Email: boutheina.yaghlane@ihec.rnu.tn}}

\maketitle

\begin{abstract}

Defining and modeling the relation of inclusion between continuous belief function may be considered as an important operation in order to study their behaviors.
Within this paper we will propose and present two forms of inclusion: The strict and the partial one.
In order to  develop this relation, we will study the case of consonant belief function.
To do so, we will simulate normal distributions allowing us to model and analyze these relations. Based on that, we will determine the parameters influencing and characterizing the two forms of inclusion.

\end{abstract}

\section{Introduction}

The theory of belief functions (also called Evidential Theory or The Dempster-Shafer Theory) is a well known framework for reasoning under uncertainty \cite{Dempster}. It is widely used to model imperfect information. We can distinguish the imprecision (lack of accuracy), the uncertainty (lack of compliance comparing to the real world, due to the source of information) and even the inconstancy where is characterized by a high level of conflict.
 
The discrete case of belief functions, knowning a large succes, have been applied to several research fields, medical, military, risk management, information fusion \cite{Martin08,Martin07,Denoeux}, etc.\\

Otherwise, Smets in \cite{BFRN}, defined the basic notions of continuous belief functions to describe them in the extended set of reals.
Recently Attiaoui {\em et al.} \cite{Attiaoui} proposed a similarity measure for continuous belief functions based on Smets' formalism using the distance of Jousselme~\cite{J1}.

Our work in this paper will consider the notion of inclusion and how two continuous belief functions can be included in each other. This operation will help us later to take into account this specific characteristic during the information fusion and the measurement of the conflit.
Thus, we will define two forms: the strict inclusion and the partial one. To do so, we will present a measure called a degree of inclusion of an interval (the focal element of a continuous belief function)  in the second one.
This work presents a new way to determine the relation of inclusion by considering a new vision for the continuous case within the theory of belief functions.

\section{Theory of belief functions: an overview}

This section recalls the necessary background related to the evidential theory. It has been developed by Dempster in his work on upper and lower probabilities \cite{Dempster}. Based on that, he was able to represent more precisely the observed data.\\
Later, in his book "A mathematical Theory of evidence'' \cite{shafer1976mathematical}, Shafer, presented that, any information defined by an expert characterized by basic belief assignments has two functions: a credibility and a plausibility function corresponding respectively to the lower and upper probabilities of Dempster.
\\
The theory was further developed by Smets in \cite{TBM,Smets89} who proposed the Transferable Belief Model (TBM). This model presents a pignistic probability induced by a belief function which is built by defining a uniform probability from each positive mass. Moreover, in terms of upper and lower probabilities, it can be considered as the center of gravity of the set of probabilities dominating the belief functions. He also introduced new tools for information fusion and decision making according to \cite{DuboisPS01}.  
\\
The objective of the evidential theory is to represent the information which is transmitted by a source concerning an event.
A belief function must take in consideration all the possible events on which a source can describe a belief. Based on that, we can define the frame of discernment.
\\

The frame of discernment is a finite set of disjoint elements noted $\Omega$ where $\Omega= \{\omega_{1},...,\omega_{n}\}$. This theory allows us to affect a mass on a set of hypotheses not only a singleton like in the probabilistic theory. Thus, we are able to represent ignorance, imprecision and uncertainty.

\begin{eqnarray}
m: 2^{\Omega} \mapsto [0,1].
\end{eqnarray}

\begin{eqnarray}
\sum_{X\subseteq \Omega} m(X) = 1.
\end{eqnarray}

The principle functions in the belief functions theory are:

\textbf{The credibility function (\emph{bel})}:  this function measures the strength of the evidence in favor of a set of propositions for all $X \in 2^{\Omega}$:

\begin{eqnarray}
  bel(X)= \sum_{Y\subseteq X, Y\neq \emptyset} m(Y).
\end{eqnarray}

The credibility is interpreted as a degree of justified support given to proposition X by the available evidence.

\textbf{The plausibility function (\emph{pl})}: expresses the maximum amount of specific support that could be given to a proposition $X \in 2^{\Omega}$. $pl(X)$ is then obtained by summing the $bba's$ given to the subsets $Y$ such that $Y \cap X \neq \emptyset$ :
\begin{eqnarray}
 pl(X)= \sum_{Y\in 2^{\Omega}, Y\cap X \neq \emptyset} m(Y).
\end{eqnarray}
\\
It measures the degree of belief committed to the propositions compatible with X. \\

\textbf{The commonality function (\emph{q}):} this function measures the set of $bbas$ affected to the focal elements included in the studied set, for all $X \in 2^{\Omega}$:

 \begin{eqnarray}
 q(X)= \sum_{Y \supseteq X} m(Y).
\end{eqnarray}

\subsection{Inclusion as a conflict measure for discrete belief functions}

Recently Martin in \cite{Martin12} defined a degree of inclusion as involved in the measurement made in order to determine the conflict during the combination of two discrete belief functions.\\
The author presented an index of inclusion having binary values where $Inc(X_{1},Y_{2})=1$ if $ X_{1} \subseteq Y_{2}$ and $0$ otherwise with $X_{1}$, $Y_{2}$ being respectively the focal elements of $m_{1}$ and $m_{2}$.
This index is then used to measure the degree of inclusion of the two mass functions and defined as:

\begin{eqnarray}
   d_{inc}= \frac{1}{|F_{1}| |F_{2}|} \sum_{X_{1}\in F_{1}} \sum_{Y_{2}\in F_{1}}  Inc(X_{1},Y_{2})
  \end{eqnarray}

   \begin{eqnarray}
   \sigma_{inc}(m_{1},m_{2})= max(d_{inc}(m_{1},m_{2}),d_{inc}(m_{2},m_{1}))
  \end{eqnarray}

Where $d_{inc}$ is the degree of inclusion  of $m_{1}$ in $m_{2}$ and inversely.
\\
This inclusion is used as a conflict measure for two mass functions, using it like presented:

\begin{eqnarray}
Conf(m_{1},m_{2})=1-\sigma_{inc}(m_{1},m_{2})d(m_{1},m_{2})
  \end{eqnarray}
where $d(m_{1},m_{2})$, is the distance of Jousselme:

\begin{eqnarray}
  d(m_{1},m_{2})=\sqrt{\frac{1}{2}(m_{1}-m_{2})^{T} \underline{\underline{D}} (m_{1}-m_{2})};
  \end{eqnarray}

 where $\underline{\underline{D}}$ is a matrix based on the dissimilarity of Jaccard expressed by $D(A,B)= 1$ if $A=B=\emptyset$  otherwise, $D(A,B)= \frac{\mid A \cap B \mid }{\mid  A \cup B \mid }$ if $\forall A,B \in 2^{\Omega}$

\section{Continuous belief functions}

In the previous sections, we have presented different specifications of discrete belief functions. Unfortunately, these functions do not allow us to manipulate continuous data that can be provided by sensors in different areas like:  search and rescue problems \cite{Dore2010}, classification issues, information fusion, etc.
\\
Some researches were interested in representing belief functions in continuous frame of discernment like Strat in \cite{Strat}, and Smets in \cite{BFRN}.
\\
In following sections, we will present the several proposals that allows us to describe continuous belief functions. First, we explain how  to extend these functions on the real numbers. To do so, we will focus on Smets' approach to represent continuous belief functions by using probability densities. Later, we will remember the other approaches for the continuous case of the theory of belief functions.

\subsection{Continuous belief functions on $\overline{\eR}$ }

Smets based on the TBM's background, used the same representation than Strat, and proposed the belief functions in the extended set of reals noted $\overline{\eR} = \eR\cup \{-\infty, +\infty\}$.\\
However, using the belief function framework to model information in a continuous frame is not an easy task mainly to the complex nature of the focal elements.
Comparing to the discrete domain, on real numbers, in (Smets 2005)  $bba$ becomes \emph{basic belief densities} $(bbd)$ defined on an interval $[a,b]$ of $\eR$.\\

\subsubsection{Basic belief densities}

A generalization of the classical $bba$ into a basic belief density ($bbd$) denoted $m^{\emph{\scriptsize I}}$ on the interval \emph{I}. He defined the $bbd$ where all focal elements are closed intervals or $\emptyset$.\\
Given a normalized $bbd$ $m^{\emph{\scriptsize I}}$, Smets defined another function $f$ on $\eR^{2}$, where:

\begin{eqnarray}
\left\{
  \begin{array}{ll}
    f(a,b)= m^{ I} ([a,b]), & a\leq b, \\
    f(a,b)=0, & a>b.
  \end{array}
\right.
\end{eqnarray}

$f$ is called a probability density function ($pdf$) on $\eR^{2}$.

\subsection{Continuous belief functions induced by probability density functions}

Let's consider several belief functions characterizing a unique source of information (the source is subjective and evidential). Smets proposed a pignistic transformation of the belief functions (representing the knowledge of the source) in order to obtain probabilities. The probabilities are used to ease the decision making. Pignistic probabilities are noted BetP having densities also noted $betf$.
For each probability density, we have a set of belief functions with which they are compatible. This set is called an \emph{Isopignitic}. The main issue is to choose one belief function from this set. To do so, we consider the principle of least commitment proposed in \cite{DPr,Hsia}. This principle supports the idea of choosing the belief function that involves the least an expert.
It can be considered as a natural approach to select the less committed \emph{bba} from a subset.
A particular type of belief functions describes the best this principle which are the consonant belief functions where focal elements are nested \cite{DPr2}.

\subsection{Continuous belief functions: other representations}

Some other approaches have been proposed in order to describe continuous belief functions.
\cite {Nguyen08} introduced in the notions of a source constituted by  a probability space  and a multivalued mapping which is able to define the lower probability defined by a $\Gamma$ function.

  This function can hold at the same time two notions: on one hand, it defines both of the lower and upper probability, on the other hand, it considers random sets.
We can say that $\Gamma$ as a multivalued mapping is measurable with respect to the spaces that it characterizes.

Moreover, he supposes that $\Gamma$ is a measurable mapping, then it is a random set by specifying its probability distribution.
Thus the probability distribution of a random set $\Gamma$ is precisely the basic probability assignment.

We say that there is a correspondence established between belief functions on a source $S$ and the probability distribution of random sets. This relation can be expressed by its density on $P(S)$. 
\\

Dor\'e {\em et al.} in \cite {Dore2011a}, proposed a similar approach founded on an index  function that can be assumed as $\Gamma$. This function can describe the set of focal elements of a continuous belief function. In this case, every index has its own probability measure where there is an allocated weight to a set of focal elements using a credal measure.
Every set is described according to its index and its probability density.

The formalism of Smets takes into consideration only to closed intervals, in \cite {Dore2010}, the author extended classical continuous belief functions by proposing belief functions where focal elements are not represented by intervals. He uses $\alpha cuts$ to measure to area of the portions of multimodal distributions.

\subsection{Similarity measure within continuous belief functions}

Attiaoui et al. in \cite{Attiaoui} proposed a similarity measure based on the distance of Jousselme using Smets' formalism. This distance uses a scalar product as a scalar product is defined on  $\overline{\eR}$ by:

\begin{eqnarray}
\langle f,g \rangle = \displaystyle{\int_{x=-\infty}^{+\infty}  \int_{y=-\infty}^{+\infty} f([x,y])g([x,y])dx dy}
\end{eqnarray}

Here, the authors presented a new method to measure the similarity  founded on the properties of belief functions on real numbers,  they were able to define a distance
between two densities in an interval $\emph{I}$.

\begin{eqnarray}
\label{CSP}
\langle f_{1},f_{2}\rangle = & &
 \int_{\tiny -\infty}^{\tiny +\infty} \!\! \int_{\tiny y_{i}=x_{i}}^{\tiny +\infty} \!\! \int_{ -\infty}^{\tiny +\infty} \int_{\tiny y_{j}=x_{j}}^{\tiny y_{j}= \tiny +\infty} \\
&& \!\!\! f_{1}(x_{i},y_{i}) f_{2}(x_{j},y_{j}) \delta(x_{i},x_{j},y_{i},y_{j})  dy_{j} dx_{j} dy_{i} dx_{i} \nonumber
\end{eqnarray}

The scalar product of the two basic belief densities is noted: $\langle f_{1},f_{2}\rangle$ with a function $\delta$ defined as $\delta: \eR \longrightarrow [0,1] $

\begin{eqnarray}
\label{IntLenEq}
\delta(x_{i},x_{j},y_{i},y_{j})= \frac{ \lambda (\llbracket max(x_{i},x_{j}), min(y_{i},y_{j})\rrbracket)}{ \lambda(\llbracket max(y_{i},y_{j}), min (x_{i},x_{j}) \rrbracket )}
\end{eqnarray}
where $\lambda$ represents the Lebesgue measure used for the interval's length and $\delta (x_{i},x_{j},y_{i},y_{j})$ is an extension of the dissimilarity of Jaccard applied for the intervals in the case of continuous belief functions.

\begin{eqnarray}
\llbracket a,b\rrbracket= \left\{
                                  \begin{array}{ll}
                                    \emptyset, & \mbox{ if }  a > b  \\
                                    $[a,b]$, & \mbox{ otherwise. }
                                  \end{array}
                           \right.
\end{eqnarray}

Therefore, the distance between two basic belief densities is defined by the following equation:

\begin{eqnarray}
\label{Dist}
d(f_{1},f_{2})= \sqrt{\frac{1}{2} (\|f_{1}\|^{2} + \|f_{2}\|^{2}-2 \langle f_{1}, f_{2}\rangle)}
\end{eqnarray}

We noticed that the standard deviation is influencing this measure. As long as the difference between the distributions grows, the more the distance is rising.
This representation proposed a natural approach that allows us to manipulate and also study the behavior of continuous belief functions induced by normal and exponential distributions.

\section{Inclusion within continuous belief functions}
Within this section, we will consider two forms of inclusion: a strict and a partial inclusion.
We will present their mathematical expressions, and explain how do we build them. But first, we enumerate several properties that must be satisfied by the relation of inclusion.

\subsection{Properties of the inclusion}

The inclusion defined between two intervals $[x_{i},y_{i}]$ and $[x_{j},y_{j}]$  in a set \textsl{I} satisfies the following requirements:

\textbf{Property 1: Non-negativity}\\

 \begin{eqnarray}
\begin{array}{l}
Inc(f_{i},f_{j})\geq 0.\nonumber
\end{array}
\end{eqnarray}

Namely, the inclusion of the first interval in the second one must never be negative.

\textbf{Property 2: asymmetry}
\\

\begin{eqnarray}
\begin{array}{l}
Inc(f_{i},f_{j}) \neq Inc(f_{j},f_{i}), \forall  f_{i}\neq f_{j}. \nonumber
\end{array}
\end{eqnarray}

No need for the  relation of inclusion to be symmetric.
\\

\textbf{Property 3: Upper bound}\\

  \begin{eqnarray}
\begin{array}{l}
  Inc(f_{i},f_{j})=1. \nonumber
\end{array}
\end{eqnarray}

  This property implies a total inclusion of the first interval in the second one other.
\\

\textbf{Property 4: Lower bound}
\\

  \begin{eqnarray}
\begin{array}{l}
  Inc(f_{i},f_{j})=0. \nonumber
\end{array}
\end{eqnarray}

This property implies the absence of any intersection or inclusion of the first  interval in the second one.

\subsection{Strict inclusion}

Here, we will define the strict inclusion between two continuous belief functions represented by two basic belief densities $bbd$.
\\
First, we use these distributions to deduce a degree of inclusion between the $bbds$ and then we can be able to measure the value inclusion between our continuous belief functions.
\\
Let's consider two continuous $pdfs$: $f_{1}$ and $f_{2}$.
If one distributions is included in the second one, then the strict inclusion is expressed by the following equation:

\begin{eqnarray}
\label{IStr}
\!\!\!\!\!\!\!\!\!\!\!\!\!\!\!\! IncStr(f_{1},f_{2})=\int_{-\infty}^{+\infty}\!\! \int_{y_{i}=x_{i}}^{+\infty} \!\! \int^{x_{i}=+\infty}_{x_{j}=-\infty}\!\! 
\int^{y_{j}=+\infty}_{y_{j}=x_{j}}   \\
 \delta_{IncStr}(x_{i},y_{i}, x_{j},y_{j}) f_{1}(x_{i},y_{i}) f_{2}(x_{j},y_{j}) dy_{j} dx_{j} dy_{i} dx_{i} \nonumber
\end{eqnarray}

Where $[x_{i},y_{i}]$, $[x_{j},y_{j}]$ are the considered intervals and $\delta_{IncStr}(x_{i},y_{i}, x_{j},y_{j}) $ is the degree of strict inclusion that will allow us to measure the value related to the inclusion of the first interval in the second one.
\\

We will consider that $\delta_{IncStr}(x_{i},y_{i}, x_{j},y_{j})$ is having binary values where:

\begin{eqnarray}
\label{IncStr}
\delta_{IncStr}(x_{i},y_{i}, x_{j},y_{j})= \left\{
                                  \begin{array}{ll}
                                    1, & \mbox{ if } [x_{i},y_{i}] \subseteq[x_{j},y_{j}]  \\
                                    0, & \mbox{ otherwise. }
                                  \end{array}
                           \right.
\end{eqnarray}

If we are in presence of two distributions that do touch each other, there is an intersection between them. The $\delta_{IncStr}(x_{i},y_{i}, x_{j},y_{j})$ will have the value $1$, and the strict inclusion will be weighted by the masses of our continuous belief functions.
Otherwise $\delta_{IncStr}(x_{i},y_{i}, x_{j},y_{j})$ will be null.

\subsection{Partial inclusion}

Considering two $bbds$ represented by two intervals $[x_{i},y_{i}]$ and $[x_{j},y_{j}]$: We say that $[x_{i},y_{i}]$ is partially included in $[x_{j},y_{j}]$ or inversely, if and only if their intersection is different of $\emptyset$.
\\
To represent the partial inclusion we define:

\begin{eqnarray}
\label{IPar}
IncPar(f_{1},f_{2})= \int_{-\infty}^{+\infty}\!\! \int_{y_{i}=x_{i}}^{+\infty} \!\! \int^{x_{}=+\infty}_{x_{i}=-\infty}\!\! \int^{y_{i}=+\infty}_{y_{j}=x_{j}} &&  \\ \delta_{IncPar}(x_{i},y_{i}, x_{j},y_{j}) f_{1}(x_{i},y_{i}) f_{2}(x_{j},y_{j}) dy_{j} dx_{j} dy_{i} dx_{i} \nonumber
\end{eqnarray}

with $\delta_{IncPar}(x_{i},y_{i}, x_{j},y_{j})$ is the degree of partial inclusion:

\begin{eqnarray}
\label{IncPar}
\!\!\!\!\delta_{IncPar}(x_{i},y_{i}, \!x_{j},y_{j})\!\!=\!\! \frac{max(0, \!min(y_{j},y_{i})\!\!-\!\!max(x_{i},x_{j}))}{y_{i}-x_{i}}
\end{eqnarray}

The degree $\delta_{IncPar}(x_{i},y_{i}, x_{j},y_{j})$ represents the length of the intersection of the two probability density functions $f_{1}$ and $f_{2}$ on the length of $f_{1}$ if we are measuring $IncPar(f_{1},f_{2})$ and the length of $f_{2}$ if we have to calculate the partial inclusion of $f_{2}$ in $f_{1}$: $IncPar(f_{2},f_{1})$

\section{Asymmetry within the inclusion}

\begin{center}
\begin{figure}
\center
\includegraphics [scale=0.5]{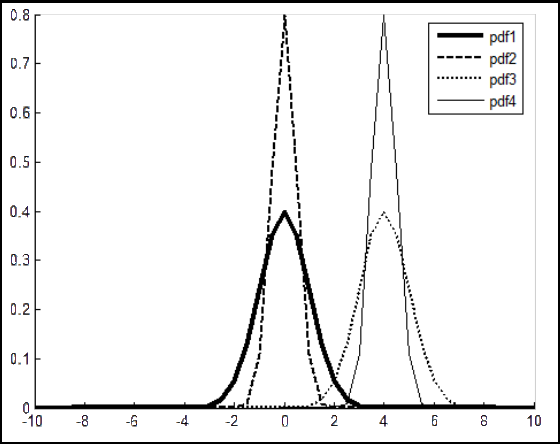}\\
\caption{Modeling probalility distribution functions. \label{pdfs}}
\end{figure}
\end{center}

Let us consider four probability density functions defined by their means $\mu$, and their standard deviations  $\sigma$, like presented in table~\ref{table1} and described in figure~\ref{pdfs}.

\begin{table}
\begin{center}
\begin{tabular}{|c|c|c|c|c|}
  \hline
  $pdf$    & 1 & 2   & 3 & 4 \\
 \hline
  $\mu$    & 0 & 0   & 4 & 4 \\
 \hline
  $\sigma$ & 1 & 0.5 & 1 & 0.5 \\
  \hline
\end{tabular}
\end{center}
\caption{Probability density distributions}
\label{table1}
\end{table}

Once we apply the mathematical formula proposed in equation (15), we obtain the following table~\ref{tableAsyStrict}  strict inclusion.

\begin{table}
\begin{center}
\begin{tabular}{|c|c|c|c|c|}
  \hline
  $IncStr$ & $pdf_{1}$     & $pdf_{2}$  & $pdf_{3}$    & $pdf_{4}$ \\
  \hline
  $pdf_{1}$  & 0.5032        & 0.1437    &   0.009   & 0 \\
  \hline
  $pdf_{2}$  & 0.8586        & 0.5032    &     0.053   &  0  \\
  \hline
  $pdf_{3}$  &     0.009     & 0        & 0.5032   & 0.143  \\
  \hline
  $pdf_{4}$  &    0.537      &    0   &    0.8586      & 0.5032 \\
  \hline
\end{tabular}
\end{center}
\caption{Strict inclusion and asymmetry.}
\label{tableAsyStrict}
\end{table}

Otherwise, applying the partial inclusion expressed in equation (17),  we obtain table~\ref{tableAsy}.

\begin{table}
\begin{center}
\begin{tabular}{|c|c|c|c|c|}
  \hline
  $IncPar$ & $pdf_{1}$     & $pdf_{2}$  & $pdf_{3}$    & $pdf_{4}$ \\
  \hline
  $pdf_{1}$  & 0.8183    & 0.5498    & 0.0253     & 0.0013 \\
  \hline
  $pdf_{2}$  & 0.9595    & 0.8183        & 0.0041     & 0.0017 \\
  \hline
  $pdf_{3}$  & 0.0253    & 0.8247        & 0.8183      & 0.9595 \\
  \hline
  $pdf_{4}$  & 0.0041    & 0.0017       & 0.5498     & 0.8183 \\
  \hline
\end{tabular}
\end{center}
\caption{Partial inclusion and asymmetry.}
\label{tableAsy}
\end{table}

The property of asymmetry between two continuous belief functions can be confirmed when we observe the measures of inclusion in the table~\ref{tableAsyStrict} and  table~\ref{tableAsy} .
\\

We witness for strict and partial inclusion that $IncStr(pdf_{2},pdf_{1})=IncStr(pdf_{3},pdf_{4})=0.8586$ and 
$IncPar(pdf_{2},pdf_{1})=IncPar(pdf_{3},pdf_{4})=0.9595$ do always have the highest values on both cases.  These primary results confirm the distributions presented in Figure~\ref{pdfs}, 

Meanwhile, for the case of inclusion of $pdf_{1}$ in $pdf_{4}$ and $pdf_{4}$ in $pdf_{1}$, where the difference between the means is very important we are dealing with very small values of inclusion.
Knowing that $\sigma_{1}= \sigma_{3}$ and $\sigma_{2} = \sigma_{4}$, when computing these partial inclusions, the difference between the standard deviations is maximal, generates a considerable value.
\\
For the strict inclusion we have some values where there is no intersection between the distributions and then we naturally obtain a null value, like for $IncStr(pdf_{1},pdf_{4})$.

According to the data in table~\ref{tableAsyStrict} and table~\ref{tableAsy}, the primary results obtained using both of the strict and the partial inclusion, our proposed relation responds to all the properties previously announced (non negativity, asymmetry, upper-bound $<1$).

\subsection{Average of inclusion}
Let us consider $n$ distributions, and $\alpha f $ a set of $bbds$.We can measure the average of inclusion of a $bbd$ ${f_{i}}$ in  $\alpha f$.
\\

To do so, we present the following equations the first one related to the strict inclusion and the second one to the partial. 

\begin{eqnarray}
 IncS(f_{i},\alpha f)=\!\!\!\!\ \displaystyle \frac{1}{n-1} \sum_{j=1, i\neq j} ^{n} IncStr(f_{i},f_{j})
\end{eqnarray}

\begin{eqnarray}
IncP(f_{i},\alpha f)=\displaystyle  \frac{1}{n-1} \sum_{j=1, i\neq j} ^{n} IncPar(f_{i},f_{j})
\end{eqnarray}

To model the measurement of the average of inclusion, we will apply the equations (19) and (20) to obtain respectively table~\ref{AvSt} and table~\ref{AvPar}

\begin{table}
\begin{center}
\begin{tabular}{|c|c|}
  \hline
  IncS & Value    \\
  \hline
  $IncS(f_{1},f_{j})$  & 0.0509      \\
  \hline
   $IncS(f_{2},f_{j})$ & 0.3038    \\
  \hline
   $IncS(f_{3},f_{j})$  & 0.0506    \\
  \hline
   $IncS(f_{4},f_{j})$  & 0.4653   \\
  \hline
\end{tabular}
\end{center}
\caption{Average of Strict Inclusion.}
\label{AvSt}
\end{table}

\begin{table}
\begin{center}
\begin{tabular}{|c|c|}
  \hline
  $IncP$ & Value     \\
  \hline
  $IncP(f_{1},f_{j})$  & 0.1921      \\
  \hline
   $IncP(f_{2},f_{j})$ & 0.3217    \\
  \hline
   $IncP(f_{3},f_{j})$  & 0.60317    \\
  \hline
   $IncP(f_{4},f_{j})$  & 0.1852 \\
  \hline
\end{tabular}
\end{center}
\caption{ Average of the Partial Inclusion.}
\label{AvPar}
\end{table}

\section{Illustration of the strict and partial inclusions}
In this section we will present our experimental phase to illustrate both of the strict and partial inclusion between two continuous belief functions.

To illustrate both of the strict and partial inclusion between two continuous belief functions induced using a normal distribution. We decided to fix the value of the first distribution $pdf_{1}$, where, it is characterized by its mean $\mu_{1}=0$ and its standard deviation $\sigma_{1}=1$. For the second distribution $pdf_{2}$, we will vary the values of $\mu_{2}$ in $[0,10]$ and $\sigma_{2}$ in $[0,5]$.
\\
Here, our purpose is to see the behavior of the inclusion when we modify one of the $pdfs$, and the parameters that infer in the obtained results.

\subsection{Strict inclusion between belief densities induced by normal distributions}

This strict inclusion is a natural approach, that allows us to perceive if there exists any intersection between two distributions, how they behave and the parameters that interfere during this process.
In this case, when a degree of inclusion has binary values  of $0$ and $1$, we will study this property between two continuous belief functions.
\\

\begin{figure}
\center
\includegraphics [scale=0.6] {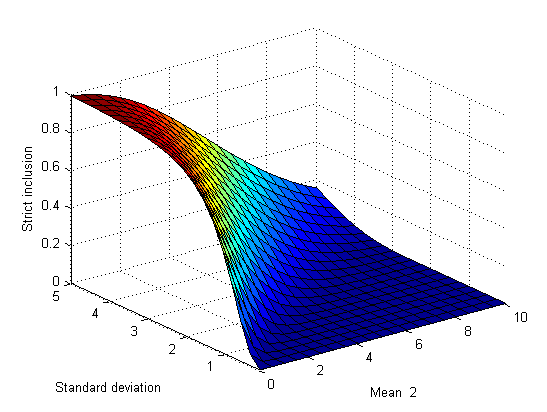}
\caption{ Strict inclusion of $pdf1$ in $pdf2$. \label{IncGlob}}
\end{figure}

The more the value of the second standard deviation grows, the more difference between $\sigma_{1}$ and $\sigma_{2}$ develops.
Here, the strict inclusion between the two distributions, which generates a growth in the behavior of the obtained curve. We are then, in the presence of a growth of the strict inclusion of $pdf_{1}$ in $pdf_{2}$ due to the variation of the second distribution.
\\
Otherwise, we notice in Figure~\ref{IncGlob}, relative to the strict inclusion, that it is not the gap between the means $\mu_{1}$ and $\mu_{2}$ that generates this growth. It is the difference between the two standard deviations, that is in the origin of this growing phenomenon of inclusion.
As shown when $\sigma_{1}>\sigma_{2}$ and $\mu_{2}$ having its maximal value with $\mu_{2}=5$, the strict inclusion is null.
This value is due to the fact that the degree of strict inclusion $\delta_{IncStr}(x_{i},y_{i}, x_{j},y_{j})=0$, which can be explained by the non-existence of any intersection between $pdf_{1}$ and $pdf_{2}$.
\\
The part of the curve where the difference between the two standard deviations is low and at the same time  the difference between the means is high. We have a small growing strict inclusion for example when $\mu_{2}=3$ and $\sigma_{2}=1.5$, the strict inclusion $IncStr(f_{1},f_{2})=0.1$.
At the meantime, even when $\mu_{1}=\mu_{2}=0$ $\sigma_{1}=\sigma_{2}=1$, there is an intersection between the two distributions and $IncStr(f_{1},f_{2})>0$.
Moreover, when the gap between $\sigma_{1}$ and $\sigma_{2}$ gets higher, the curve increases, generating a bigger strict inclusion between $pdf_{1}$ and $pdf_{2}$, until rising its maximal value where $IncStr(f_{1},f_{2})=1$, with $\delta_{IncStr}(x_{i},y_{i}, x_{j},y_{j})=1$.
In this specific case, we are in presence of a total inclusion of the first distribution $pdf_{1}$ in the second one $pdf_{2}$.

We also witness a phenomenonin which $\mu_{2}$ gets a high value, the strict inclusion drops considerably. Here we can state that the mean also has an impact on the generated inclusion.

\subsection{Partial inclusion between belief densities induced by normal distributions}

The partial inclusion is defined in order to give us the proportion of the intersection between two $pdfs$.

\subsubsection{Partial inclusion of $pdf1$ in $pdf2$}

\begin{figure}
\center
\includegraphics [scale=0.6] {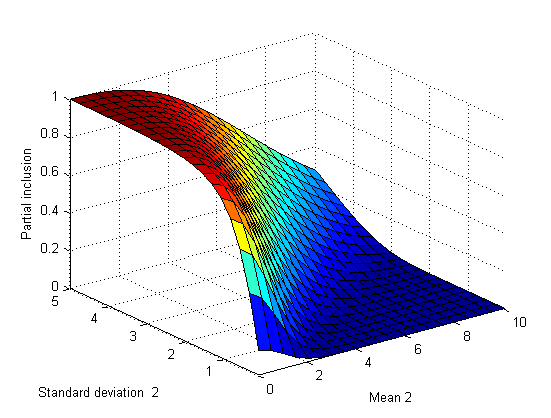}
\caption{Partial inclusion of $pdf1$ in $pdf2$ \label{IncPart}}
\end{figure}

During this experimentation, we keep the same values used for the strict inclusion and we obtain Figure~\ref{IncPart}
We notice that, when we are dealing with similar distribution where $\mu_{1}=\mu_{2}=0$ and $\sigma_{1}=\sigma_{2}=1$, the value of the partial inclusion is greater than zero. It is possible to state that when we are in presence of two distributions having the same values, there is not necessarily any total inclusion between them.
We also take note, that as the difference between $\sigma_{1}$ and $\sigma_{2}$ rises, due to the  of $pdf_{2}$, the figure obtained grows faster, and reaches its maximal value $IncPar(f_{1},f_{2})=1$, generating a curve more arched than the strict inclusion.

When the difference between $\mu_{1}$ and $\mu_{2}$ increases because of the variation of $pdf_{2}$, the partial inclusion reaches a value of $IncPar(f_{1},f_{2})=0.85$, which becomes lower when the gap between two standard deviations is the highest ($\sigma_{2}=3$), we obtain the maximal value for the partial inclusion: $IncPar(f_{1},f_{2})=1$ like presented in Figure~\ref{IncPart}.\\
In this specific case, we witness a full and total inclusion of $pdf_{1}$ in $pdf_{2}$. This is similar to what we have presented regarding the strict inclusion.
Here we have non-negative inclusions, that respect the lower and upper bounds where the values are $[0,1]$.
Besides, the property of asymmetry is also respected because, the inclusion (what ever is strict or partial) of a distribution in the other does not necessarily involve the inverse case with the same value.

When $IncPar=1$, we have a total inclusion of the first distribution $pdf_{1}$ in the second one $pdf_{2}$, this situation is considered as a strict inclusion where, $pdf_{1}$ is fully included in $pdf_{2}$.

\subsubsection{Partial inclusion of $pdf2$ in $pdf1$}

\begin{figure}
\center
\includegraphics [scale=0.6] {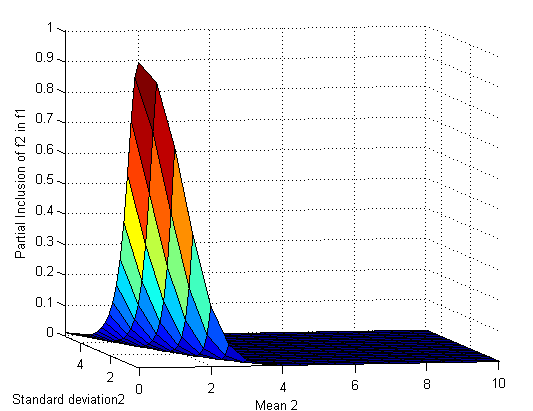}
\caption{Partial inclusion of $pdf2$ in $pdf1$ \label{Incp2}}
\end{figure}

For this case, we have chosen to measure the partial inclusion of $pdf_{2}$ in $pdf_{1}$, saving the same values for both distributions. Otherwise, the equation 18 becomes:

\begin{eqnarray}
\!\!\!\!\!\!\!\! \delta_{IncPar}(x_{i},y_{i}, \! x_{j},y_{j})\!\!\!=\!\! \frac{max(0, \! min(y_{i},y_{j}) max(x_{j},x_{i}))\!\!\!}{y_{j}-x_{j}}
\end{eqnarray}

We obtain the Figure~\ref{Incp2}. In this case, we notice a different phenomenon comparing to Figure~\ref{IncPart}.
Here, we witness than when the two distributions are totally similar, the value of the partial inclusion equals $0.8183$. This value is the maximal one that is expressed in this, and as stated before, two similar distributions can not be fully included in each other.
We also observe in this figure, that as long as tha values of the second distribution $\mu_{2}$, and $\sigma_{2}$ rise, the value of the partial inclusion of $pdf_{2}$ in $pdf_{1}$ drops. 
More the two distributions are getting different from each other, more the partial inclusion decreases. Both parameters; the mean and the standard deviation have considerable impacts in this measure. This is proved by the behavior of the partial inclusion.
\\

The difference between the Figure~\ref{IncPart} and Figure~\ref{Incp2} is very obvious considering the behaviors of the two curves.
This is due to the nature of the focal elements of continuous belief functions which are expressed by intervals.
Specially with the case of the inclusion where each time we measure the inclusion of the intervals of a normal distribution with those of a second normal distribution.

\section{Strict inclusion VS Partial inclusion } 
Comparing the results obtained in Figure~\ref{IncGlob} and Figure~\ref{IncPart}, where we have the same values for the two distributions, we notice that the partial inclusion reaches the maximal value faster that the strict one. Besides, its area is bigger and larger. Thus, we can state that the partial inclusion is dominating.
\\
In Figure~\ref{IncGlob} and Figure~\ref{IncPart}, it seems that we have the same phenomena between the two types of inclusion. This can be explained by the fact that we are working with consonant belief functions (where focal intervals are nested).
We can imagine that in presence in other nature of belief functions (categorical, or Bayesian belief functions) we can obtain different behaviors between the strict and the partial inclusion.

We notice the same phenomena that we have seen for the strict, when the values of parameters characterizing the second distribution grow, generating a big difference between the means, and the standard deviations, the partial inclusion decreases significantly and comparing when we are dealing with smaller values of $\mu_{2}$.

Thus, we can say that, both of the mean and the standard deviation do have a real impact on the measurement of the partial inclusion, having the same situation as the strict inclusion.

In the case of two distributions getting more and more different and especially when considering only the standard deviation, the phenomena of inclusion gets bigger. However, if the mean of one distribution is having non similar value than the other we can state than the inclusion has smaller values.

\section{Conclusion}
In this paper, we have emphasized the evaluation of the relation of inclusion between continuous belief functions induced by a normal distribution. We have defined two forms of inclusions: the strict and the partial one.
Before that, we have detailed all preliminary background that will allow us to experiment this kind of relation. We have also provided the approach on which the evaluation of the inclusions will be based.

We have presented a relation of inclusion having normalized values that takes into account the nature of continuous belief functions described using intervals as focal elements.
These two forms of inclusion respond to all the properties that must be satisfied.  
We also have shown that both of the mean and the standard deviation have different impacts in this phenomena each one with its specific output.

\bibliographystyle{IEEEtran}
\bibliography{IEEEabrv,mybibfile}

\end{document}